\def\year{2022}\relax
\documentclass[letterpaper]{article} 
\usepackage{aaai22}  
\usepackage{times}  
\usepackage{helvet}  
\usepackage{courier}  
\usepackage[hyphens]{url}  
\usepackage{graphicx} 
\usepackage{natbib}  
\usepackage{caption} 
\DeclareCaptionStyle{ruled}{labelfont=normalfont,labelsep=colon,strut=off} 
\usepackage{algorithm}
\usepackage{algpseudocode}
\usepackage{amssymb}
\usepackage{amsmath}
\usepackage{xcolor}
\usepackage{amsthm}
\usepackage{adjustbox}
\usepackage{multirow}

\ifx\state\undefined
\usepackage[final]{changes}
\else
\usepackage[markup=underlined]{changes}
\usepackage[marginparwidth=2.5cm]{geometry}
\fi

\usepackage{todonotes}

\usepackage{enumitem}
\newlist{myitemize}{itemize}{1}
\setlist[myitemize,1]{label=\textbullet,leftmargin=0pt}

\newtheorem{definition}{Definition}

\usepackage{newfloat}
\usepackage{listings}
\floatstyle{ruled}
\newfloat{listing}{tb}{lst}{}
\floatname{listing}{Listing} 
\copyrighttext{Presented at the AI-HRI Symposium at AAAI Fall Symposium Series (FSS) 2022}
\title
{
Robust Planning for Human-Robot Joint Tasks \\ with Explicit Reasoning on Human Mental State
}
\author{
    Anthony Favier\textsuperscript{\rm 1,2},
    Shashank Shekhar\textsuperscript{\rm 1},
    Rachid Alami\textsuperscript{\rm 1,2}
}
\affiliations{
    \textsuperscript{\rm 1}LAAS-CNRS, Universite de Toulouse, Toulouse, France\\
    \textsuperscript{\rm 2}{Artificial and Natural Intelligence Toulouse Institute (ANITI)}
    \{anthony.favier, sshekhar, rachid.alami\}@laas.fr
}

\begin{document}

\newcommand{\worldstates}{\mathcal{S}}
\newcommand{\worldstate}{s}
\newcommand{\fluent}[3]{\mathcal{F}^{#1#2}_{#3}}
\newcommand{\prop}{\varphi}
\newcommand{\allprops}{\Phi}
\newcommand{\predicate}{\mathcal{P}}
\newcommand{\predval}{v}
\newcommand{\agent}{\lambda}
\newcommand{\beliefs}{\mathcal{B}}
\newcommand{\human}{H}
\newcommand{\robot}{R}
\newcommand{\places}{\mathcal{P}}
\newcommand{\place}{p}
\newcommand{\unknown}{UnKw}
\newcommand{\known}{Kw}
\newcommand{\missedactions}{\mathcal{M}}
\newcommand{\loc}[1]{loc(#1)}
\newcommand{\obs}[1]{obs(#1)}
\newcommand{\dom}[1]{dom(#1)}
\newcommand{\observable}{\texttt{OBS}}
\newcommand{\inferrable}{\texttt{INF}}

\maketitle

\begin{abstract}
We consider the human-aware task planning problem where a human-robot team is given a shared task with a known objective to achieve. Recent approaches tackle it by modeling it as a team of independent, rational agents, where the robot plans for both agents' (shared) tasks. However, the robot knows that humans cannot be administered like artificial agents, so it emulates and predicts the human's decisions, actions, and reactions. 
Based on earlier approaches,
we describe a novel approach to solve such problems, which models and uses {\em execution-time observability conventions}. 
Abstractly, this modeling is based on situation assessment, which helps our approach capture the evolution of individual agents' beliefs and anticipate belief divergences that arise in practice.
It decides {\em if} and {\em when} belief alignment is needed and achieves it with communication.
These changes improve the solver's performance: (a) communication is effectively used, and (b) robust for more realistic and challenging problems.
\end{abstract}

\section{Introduction}

Consider a scenario where you (the \textit{human} agent) want to prepare, together with a robot, some pasta in your kitchen. This joint task comprises several \textit{components} and \textit{activities}, e.g., pasta is kept either in the kitchen or the living room, covering a pot, turning a furnace on, 
the salt container, adding salt to the pasta, etc. 
Sometimes, some components and activities might be accessible only to you (the robot) from your (its) current position, while some are accessible to both. 
Like this subclass of joint task, its multiple instantiations exist across many real-world domains, e.g., homes, offices, hospitals, etc.

Naturally, while achieving such tasks, you would want a pleasant and flexible collaboration. This involves behaviors like not being bothered too much by the robot or being able to delegate tasks to it, even if it delays the overall process.
Here, we consider the human agent cooperative and rational but cannot be administered like a robot (a \textit{controllable} agent). 
So, if the human agent has several ways to achieve a goal or accomplish a task, the robot cannot instruct them explicitly which one to choose. However, it can still act to influence their choices, thus eliciting future actions. 

The above example outlines an important point about human-robot collaboration. The robot should not only plan for agents' joint actions but also predict and emulate human decisions, actions, and reactions, to achieve a joint goal seamlessly. 
It must put itself in the shoes of the human in order to better predict their decisions based on an accurate estimation of their beliefs; this would open the possibility for the robot to “create” circumstances that promote actions to be performed by humans and relevant to achieving the joint task or to prevent them from making errors based on false, outdated or inaccurate beliefs.


Human-aware planning has been a topic of interest, and a  framework proposed and upgraded over the years is called HATP (Human-Aware Task Planner)~\cite{alami2006toward,alili2009task,lallement2014hatp,lallement2018hatp}. Precisely, the HATP framework extends the existing Hierarchical Task Network (HTN) framework~\cite{naubooks0014222}, extending its representation and semantics to make it more suitable for robotics and, more specifically, human-robot collaboration. The agents become the first class entities while ``social rules'' are added to classify planned behaviors for the robot as (un)acceptable.

A more elaborate planning architecture, inspired by HATP although extends HTNs differently, has been recently proposed~\cite{BuisanA21,buisan:hal-03684211}.
It introduced the \textit{first} planning framework of this type, discussing the abilities and framework's broad spectrum.
It is called HATP/EHDA, which extends HATP by \textbf{e}mulating \textbf{h}uman \textbf{d}ecisions and \textbf{a}ctions. 
The new planning scheme raises several interesting, non-trivial questions to be investigated, and their principled will improve the proposed HATP/EHDA framework toward task-level autonomy for human-robot interaction or human-robot collaboration.

HATP/EHDA comprises a \textit{dual}-HTN structure that can be seen as a joint hierarchical task specification model. 
We assume such specifications are given by a domain modeler. The joint model describes agents' capabilities, initial beliefs, shared tasks, world dynamics, understanding of common ground, etc. Moreover, the modeler implicitly captures hypothetical variables to represent the human mood, intentions, etc., that are non-trivial to manage explicitly but ``affect'' the human behavior. 

HATP allows only one search thread for a shared plan, resulting in two coordinated plan streams for the robot and human team. 
However, HATP/EHDA proposed a two-threaded search process, generating a policy tree comprising both agents' actions. 
The human agent is \textit{uncontrollable}, so it is not trivial to determine their exact ``next'' feasible action for the overall joint task. It captures the impact of their mental state, mood, intention, etc.  
Also, it is important to note that both the robot's and human's HTNs are within the robot. At each stage, it must plan their joint actions following the specifications while predicting, managing, and emulating human's decisions, goals, beliefs, etc.

However, the current implementation of the HATP/EHDA framework
makes some simplistic assumptions. E.g., an action performed by an agent would also impact the beliefs of other agents, i.e., considering that other agents are always aware of the action execution and share the same perspective as the actor. 
In reality, a robot's action can influence human's beliefs differently under different scenarios. For example, adding salt to the pasta or switching on the furnace (in the kitchen) would impact the human's belief state differently depending on whether the human is also in the kitchen or not at the time of action execution.   



We introduce a novel paradigm for implementing \textit{suitable conventions} understood during plan execution in this context. This paradigm helps upgrade the HATP/EHDA's planning system. 
A high-level idea to formalize it is inspired by existing situation assessment reasoners which perform spatial reasoning and perspective taking~\cite{flavell1992perspectives,trafton2005enabling,johnson2005perceptual,Sisbot2011SituationAF,warnier-2012,lemaignan-2017}. 
The upgraded planning system handles divergences in human-robot individual beliefs (computed using our proposed model) and plans with explicitly modeled communication actions such that it tackles belief divergences via communication, if and when needed.


Effective use of a communication modality is essential to achieve the motivation behind their development, say, a seamless collaboration. However, in reality, communication incurs some costs. 
Moreover, in human-robot interaction (HRI), communicating too much or too little can disturb the overall task achievement process. 
Hence, deciding ``how'' to communicate is crucial, but it is equally crucial to decide, ``if'', ``when'', and ``what'' to communicate. 
Assume that a \textit{speech} modality is available while the 
\textit{decisional} aspect of communication is handled by the new solver with an enhanced reasoning system on the human mental state.
%


\section{Related Work}

Several attempts to model human activities are made such that these models are used to design a system, which integrates into another paradigm dealing with human tasks to improve the latter's performance. 

A common way to represent human activity and interaction with a computer at an abstraction level is by using \textit{task models}. The hierarchical structure of activities was first exploited by Annett and Duncan (\citeyear{annett1967task}), and they said a task could be seen at several abstraction levels until some criterion is satisfied. Each can, thus, refine into more concrete subtasks detailing the procedure followed by the human to achieve the higher-level task.
Task modeling evolved to introduce system interactions, produced and needed information, potential errors, and a variety of operators specifying how tasks interact with each other during their execution. Task models' usage is common in user-centered and user interface designing processes. Most advanced notations include {\sc ConcurTaskTrees}~\cite{paterno2004concurtasktrees} and {\sc hamsters}~\cite{martinie2019analysing}.
Such models are used for designing or evaluating interactive systems, provide a task understanding, and inspire this work since they share a high-level architecture with HTNs. 

Previous approaches are based on decomposing tasks hierarchically to human-aware task planning and assume a fully controllable and cooperative human agent willing to participate in achieving a  joint-goal~\cite{alami2006toward,alili2009planning,lallement2014hatp,de2015hatp,lallement2018hatp}. 
Moreover, the agents are assumed to have established a shared goal before planning. 
Later, the generated plan gets shared with the human before the execution~\cite{milliez2016using}.
We note that HATP does not represent humans as \textit{regular} agents with separate decision processes, which may lead to diverging plans without robot communication.

Other approaches are distantly related to what we do, consider an external human model (e.g., Agent Markov Models (AMMs)~\cite{unhelkar2020decision,UnhelkarLS19}), which predicts human activities, and hence the robot plans accordingly~\cite{hoffman2007effects,unhelkar2020decision,UnhelkarLS19}. Such systems can determine actions that influence human future actions. While some systems interact with humans even when they are non-collaborating~\cite{buckingham2020robot}.

In human psychology, especially in human-human interaction or collaboration, spatial reasoning and perspective-taking are crucial~\cite{flavell1992perspectives,tversky1999speakers}. Through them, a person can mimic the mind of another person to understand their point of view. Indeed, Theory of Mind (ToM) refers to the ability to ascribe distinct mental states to other people and update them by reasoning about their perceptions and goals~\cite{premack1978does,baron1985does}.

Ideas from psychology got employed in many human-robot contexts. In
robotics, ToM often
focuses on perspective taking and belief management such that a robot reasons out what humans can perceive~\cite{berlin2006perspective,milliez2014framework} and builds representations of the environment from their perspective, which sometimes helps solve ambiguous tasks, predict human behavior, etc.
\citeauthor{johnson2005perceptual} (\citeyear{johnson2005perceptual}) propose a method based on visual perspective taking for a robot recognizing an action executed by another robot. 
While~\citeauthor{milliez2014framework} (\citeyear{milliez2014framework}) propose visual and spatial perspective taking to find out {\em the referent} indicated by the human. 
In~\cite{Sisbot2011SituationAF}, based on spatial reasoning and perspective taking, a reasoner generates online (symbolic) relations between agents and objects {\em co-present} in the environment. 
Such relations are stored in databases, allowing access to the complete framework, and used for planning, acting, or both.

The framework proposed in~\cite{devin2016implemented} allows the robot to estimate the mental state of the human, which contains not only the human's belief but their actions, goals, and plans. It supports the robot's capabilities to do spatial reasoning w.r.t. the human and track their activities. In particular, it shows how to manage the execution of shared plans in a collaborative object manipulation context and how a robot can adapt to human decisions and actions while communicating when needed.

This work is motivated by the main ideas of ToM~\cite{devin2016implemented} and situation assessment based on spatial reasoning and observability while abstractly employing them during planning. 

Communication is a key to successful human-robot collaboration, used to align an agent's belief, clarify its decision or action, fix errors, etc.~\cite{tellex2014asking,sebastiani2017dealing}. We already discussed the ToM-enabled framework by Devin and Alami (2016). The framework handles execution time subtleties such that it decides at the execution time if communication is needed and then the content that should be transmitted. 
They achieve it by monitoring the divergence between the robot knowledge and the estimated human knowledge. If a belief divergence is detected that can endanger the plan, then verbal communication takes place. 
Recent work deals with an explicit usage of communication actions in planning~\cite{BuisanSA20,nikolaidis2018planning,roncone2017transparent,sanelli2017short,UnhelkarLS20}. 
E.g., in~\cite{roncone2017transparent,UnhelkarLS20}, the authors represent and plan with explicit communication actions, considering them as regular POMDP actions, such that execution policies generated contain communication actions.

However, this work estimates the evolution of the agents' beliefs and decides ``if'' and ``when'' \textit{belief alignment} is necessary for planning. 
And, it is achieved via explicit communication actions, answering ``what" to communicate. 
Our solver does not use these actions (for the deliberation process) like (non-) primitive tasks but to minimally align an agent's belief with the ground reality so that the belief divergence does not impact the overall task achievement. 
At each stage, if needed, a \textit{sequence} of communication actions is \textit{computed} by a modified Breadth-First Search planning subroutine and \textit{appended} to the sender's plan. We provide clarification on this when we formalize our contributions.

\section{Relevant Background}

\subsection{Hierarchical Task Networks}
Generalized basic terminologies and definitions related to HTNs are presented, e.g., task network, problem, and solution definitions~\cite{naubooks0014222}.  

A \textit{task network} is a 2-tuple $w=(U,C)$, where $u\in U$ is a task node, while $t_u$ is the task associated to this node. $C$ is the set of constraints - including orderings between task nodes, binding constraints, etc. If $\forall u \in U$, $t_u$ is a primitive task, the task network is primitive; otherwise, non-primitive.

\begin{definition}
\textbf{(HTN Planning Problem.)} 
The HTN planning
problem is a 3-tuple $\mathcal{P} = (s_0, w_0, D)$ where $s_0$ is the initial belief state (the ground truth), $w_0$ is the initial task network, and $D$ is the HTN planning domain which consists of a set of tasks and methods.
\end{definition}
A \textit{domain} is a 2-tuple $D=(O, M)$ where $O$ is the set of operators and $M$ is the set of methods. 
An operator $o \in O$ is a primitive task described as $o=(name(o), pre(o), \textit{eff}(o))$, i.e., its \textit{name}, \textit{precondition}, and \textit{effect}, respectively. 

A \textit{task} consists of a task symbol and a list of parameters. It is called a \textit{primitive} task if its task symbol is an operator name and its parameters match, otherwise, it is a \textit{non-primitive} task. A primitive task is assumed to be directly \textit{executable}, while to execute a non-primitive task, we need to decompose it into sub-tasks using \textit{methods}. 

A method ($m \in M$) is a 4-tuple $m=(name(m),task(m),subtasks(m),constr(m))$ which are method name, a non-primitive task, a set of sub-tasks, and a set of constraints, respectively. $m$ is relevant for a task $t$, if for a parameter substitution $\sigma$, $\sigma(t) = task(m)$. There could be different methods relevant for a single task $t$, decomposing it differently, depending on the context. $(subtasks(m),constr(m))$ is a task network in $m$.      

Suppose $m$ is an instantiated method, and $task(m)=t_u$ then $m$ decomposes $u$ into $subtasks(m')$, producing a new task network, $\delta(w,u,m)=((U-\{u\})\cup subtasks(m'),C'\cup constr(m'))$.
Here, every \textit{precedence}, \textit{before}, \textit{after}, and \textit{between} constraints between tasks are carefully maintained after each decomposition.

\begin{definition}   
(\textbf{Solution Plan}.) 
{A sequence of primitive actions $\pi=(o_1,o_2,o_3...,o_k)$ is a solution plan for the HTN planning problem $\mathcal{P}=(s_0,w_0,D)$ iff there exists a primitive decomposition $w_p$ (of the initial task network $w_0$), and $\pi$ is an instance of it. 
}  
\label{def:htn-sol-plan}
\end{definition}

\subsection{The HATP/EHDA Framework}
The framework comprises a dual-HTN based joint-task specification model.   
For ease of exposition, consider a team of a single human and a single robot. Two categories: the robot model, $\mathrm{HTN}_{r}$, represents the task specifications for the controllable agent, while $\mathrm{HTN}_{h}$, represents the model for the human, who is an uncontrollable one but the planner has its model
representation and relevant interpretations to predict and emulate their decisions, actions, and reactions. 


We restrict the framework's description to what is relevant but briefly discuss its ability to capture a broad class of scenarios. It supports the robot to act in the presence of a human agent even when they do not share a task to achieve in the beginning. Or, it can also support the robot planning for both agents by asking the human to help it occasionally, can manage the creation of shared tasks, handle human reactions modeled explicitly via triggers, etc. More details on triggers  in~\cite{ingrand1996prs,AlamiCFGI98}.   

A brief working description of the HATP/EHDA framework is as follows. 
The structure manipulated by it is ``agent''~\cite{thesisBuisan21,buisan:hal-03684211}.
Two such structures are used, one for the \textit{human} and one for the \textit{robot}, each includes a model of the corresponding agent.
Each model (structure) has its own belief state, action model, task network, plan, and triggers. We are interested in solving the HATP/EHDA problem, $\mathcal{P}_{hr}$, in which agents start with a joint task, i.e., a shared initial task network, $w_0$, to decompose.
The existing planner uses agents' action models and beliefs to decompose the given task network into its legal primitive components. 
Decomposition updates the current network by inserting new (non-) primitive tasks, additional constraints, etc., such that the single-agent process is generalized for the two-agent scenario.
While doing so, it also updates the belief state of each agent and models their reaction by executing the triggers.



A simplifying assumption is that agents \textit{act} alternatively, while only one is \textit{allowed} to act at a time.
Hence, the framework may face problems dealing with real concurrent actions~\cite{CrosbyJR14}, especially when they are \textit{interacting}~\cite{ShekharB20}. 
As a result, the framework is sound and complete only for the problem specifications in which action concurrency is handled carefully. 
Later, one can find a sound \textit{parallelization} of the agent's plans (by respecting the flexibility proposed for the human agent) in the post-processing phase, e.g., by managing the causal links and synchronizing agents' action orderings. 
Note that it is not trivial to formally handle interacting actions in the current HATP/EHDA framework, and out of the scope of this work, and it is a part of our future work.  

First, the framework builds the whole search space considering all possible, feasible decompositions~\cite{buisan:hal-03684211}. 
The framework uses specific actions during the search.
{\em IDLE} is inserted into the agent's plan when its task network is empty or fully decomposed, and {\em WAIT}, when none of its actions is applicable.
Then, it can adapt off-the-shelf graph search algorithms, e.g., the well-known algorithms like $A^*$ and $AO^*$, and consider social cost, plan legibility, acceptability~\cite{alili2009task}, etc., to search for a \textit{joint solution plan} for the agents. 

A joint solution plan generated, prior to the preprocessing stage, is defined as follows (extending Definition~\ref{def:htn-sol-plan} for the HATP/EHDA case).


\begin{definition}
(\textbf{Joint Solution Plan}.) 
It is a solution for $\mathcal{P}_{hr}$, represented as a tree or a graph, i.e., $G=(V,E)$. Each vertex ($v \in V$) represents {\em the ground truth} starting from the root node (i.e., $s_0^r$). 
Each edge ($e \in E$) represents a primitive action performed either by the robot ($o^r$) or the human ($o^h$). 
$G$ is branched on possible human choices ($o^{h}_1$, $o^{h}_2$, ..., $o^{h}_m$). 
\label{def:hatp-joint-sol-plan}
\end{definition}

The requirement from this solution tree is that for each branch, from the root to a leaf node, the sequence of primitive actions, say, $\pi=(o^r_1, o^h_2,o^r_3,...,o^h_{k-1},o^r_k)$ is a solution plan for $\mathcal{P}_{hr}$. Here, each $o_i^h$ represents a choice (often out of several choices) the human could make during execution.


\subsection{Situation Assessment} 
In human psychology, especially in human-human interaction, spatial reasoning and perspective-taking are crucial. 
In practice, we often build relations between 
entities in the environment and estimate the knowledge and capabilities of agents around us to have a seamless collaboration with them.



This work abstractly employs the main ideas of situation assessment based on spatial reasoning and observability during planning. The goal is to predict the situation assessment of a collaborative human agent to have a better estimate of their knowledge and next possible actions, and thus, produce better plans.







\section{Execution-Time Observability Conventions}
We formally model run-time observability conventions, understood during plan execution, and use them for task planning in human-robot collaboration. 

At a high-level, the formalization below is based on the standard state-variable representation (we refer the reader to~\cite{naubooks0014222}) such that we abuse the standard notations slightly, sometimes to maintain the flow of the discussion.

\subsection{A General Understanding of Common Ground}
To describe an environment a set of \textit{constant symbols} is used. 
Constants are often classified as different groups ($gr$), e.g., \textit{places}, \textit{pots}, \textit{agents}, \textit{etc.} 
Each constant can be represented as an \textit{object symbol}, e.g., \textit{robot1}, \textit{human2}, \textit{pasta-pkt1}, etc. 
An \textit{object variable} belonging to a group can be \textit{instantiated} using any constant symbol of that group.

Suppose $\mathcal{S}$ is the complete state-space and $s \in \mathcal{S}$ represents a single state (or a \textit{belief} state in the current context).  


\begin{definition}
\textbf{(State Variable Function.)} It is represented as: $f_{svs}:(?g_1 (gr_1), ?g_2 (gr_2), ..., ?g_k (gr_k),\mathcal{S})\rightarrow ?g_{k+1} (gr_{k+1})$. 
\end{definition}
Here, $svs$ is \textit{state-variable symbol} (or the attribute name), while each \textit{term} with ``$?$'' (a question mark symbol), can be instantiated with a constant symbol from its respective \textit{group} (mentioned inside ``$()$'') appears right besides the term. 
For example, to model agents' location we use $f_{\textit{AgtAt}}:(?a (Agents), \mathcal{S}) \rightarrow ?r (Rooms)$, representing, for each given legal state ($s_i \in \mathcal{S}$), the room where each agent is located. 
Each such possible \textit{instantiation} of defined state variable functions represents $s_i$ partially, and is also called a characteristic attribute of $s_i$.     

If a variable is of the form $f_{svs_i}: (?g1 (gr_1), \mathcal{S}) \rightarrow ?g_2 (gr_2)$ such that $gr_1$ contains \textit{only} one element, then, for a given state $s \in \mathcal{S}$, we simplify this expression to, $f_{svs_i}^{s} \rightarrow ?g_2 (gr_2)$. 

Suppose $s \in \mathcal{S}$ is the real state with the ground truth. As per our assumptions, the belief state of the robot w.r.t. the real-world is always the real world state, i.e., $B_{\varphi_r}^s = s$, and the human belief w.r.t. $s$ is $B_{\varphi_h}^s$ --- that is estimated when the robot takes the human's perspective. Each state variable function instantiation w.r.t. a belief state, $B_{\varphi_r}^s$ ($B_{\varphi_h}^s$), represents the truth value of that state attribute 
(grounded state-variable function)
with the \textit{perspective} of 
$\varphi_r$ ($\varphi_h$). While both these perspectives are managed by the robot.

\begin{definition}
\textbf{(Belief Divergence.)}
Suppose that $f_{\textit{svs}_i}:(g_{(1)l},g_{(2)m},...,g_{(k)n},s) \rightarrow g_{(k+1)p}$, for a legal state $s$ and $g_{(i)}$ is the $i^{th}$ group while $g_{(i)j}$ is its $j^{th}$ element, represents a possible {\em grounding} for a state-variable function.
Belief divergence is caused if there exists an instantiation such that $f_{\textit{svs}_i}:(g_{(1)l},g_{(2)m},...,g_{(k)n},B_{\varphi_r}) = {g_{(k+1)p}}  \neq f_{\textit{svs}_i}:(g_{(1)l},g_{(2)m},...,g_{(k)n},B_{\varphi_h})$.
\end{definition} 

Uncertainty in agents' knowledge is not captured explicitly, implying that they are convinced about their beliefs.

\subsection{Place-Based Observability}
We now define place-based observability criteria \textit{inspired by} some standard execution time observability conventions that are relevant for the human-robot context. 

\begin{definition}
\textbf{(Places.)} Represented as $\mathit{Places}$, 
it
captures a group of constant symbols such that each member individually captures a pre-specified area in an environment declared by the domain modeler.  
\end{definition}
Agents are always situated in a place or moving between two places. Two agents situated in a same place $p \in \mathit{Places}$, at a given time instance, are said to be \textit{co-present}.

\begin{definition}
    \textbf{(Place Specific State Attribute Function.)} A 
    {\em grounded} 
    attribute of a given state is explicitly associated with some {\em place}
    or, in a generalized way, it can be expressed as, 
    $f_{loc}: (f_{svs_i}(?g_1(gr_1),...,?g_k(gr_k),\mathcal{S})) \rightarrow Places$
    \label{def:pssav}
\end{definition}
Here, $\mathit{loc}$ is a location specific symbol for {\em place specific attribute}, while the function captures the associated place w.r.t. a state attribute, $f_{svs_i}(...)$.

Defining it this way captures and maintains the explicit places \textit{linked} with ``only'' the state attributes of our interests. 
These explicit mappings suggest that an attribute is effective (or \textit{observable}) in its dedicate \textit{place}.
Such mappings can change when an action changes the \textit{place} of effectiveness of a state attribute, e.g., 
\textit{holding} a cup and \textit{moving} to an adjacent room. Such updates are currently manually handled.


\begin{definition}
\textbf{(State Variable Observability Function.)} 
The state variable observability function maps each grounded state attribute 
to either $\observable$ or $\inferrable$; or a more general and formal way to express it is, $f_{\textit{obs}}: (f_{svs_i}(?g_1(gr_1),...,?g_k(gr_k),\mathcal{S}
)) \rightarrow 
    \{\inferrable, \observable\}$.
    \label{def:svof}
\end{definition}
Here, \textit{obs} represents an \textit{observability} symbol to capture \textit{state-variable observability} such that a state-variable function, for a legal state $s\in\mathcal{S}$, is classified as either \textit{observable} ($\observable$) or \textit{inferrable} ($\inferrable$). 
Here, we further generalize it, assuming that an attribute is either \textit{observable} or \textit{inferrable}, and hence, relaxing the individual state based restrictions, that means, the above expression can be further simplified to, $f_{\textit{obs}}: (f_{svs_i}(?g_1(gr_1),...,?g_k(gr_k))) \rightarrow 
    \{\inferrable, \observable\}$.


When a state-variable function ($f_{svs_i}(...)$) belongs to the class $\observable$, it indicates that an agent can assess/observe the correct knowledge of its \textit{exact} status in a given state $s_l$, if the agent fulfills certain requirements w.r.t. the state $s_l$. 
But when a state-variable function ($f_{svs_j}(...)$) belongs to the class $\inferrable$, unlike the previous case, it means that an agent \textit{cannot} observe it directly. However, under certain scenarios, the agent can definitely predict or infer its \textit{exact} status in $s_l$. 

\begin{figure}
    \centering
    \includegraphics[width=0.9\linewidth]{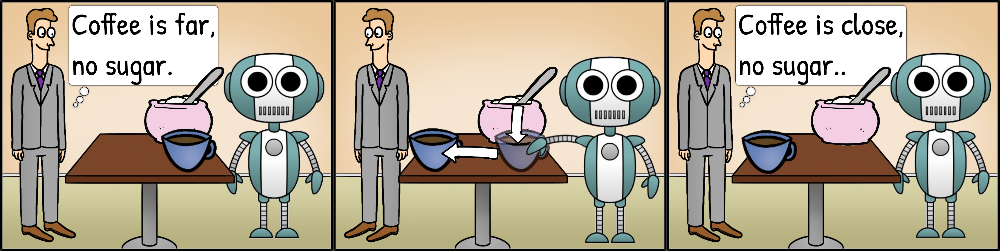}
    \caption{
    Human is not attentive when Robot adds sugar to the cup before pushing it forward. \textit{Assessing} the new situation after the two actions, Human can only ``know'' the cup's new position (\textit{observable}), but not \textit{SugarInCup} (\textit{inferrable}).
    }
    \label{fig:obs_attr}
\end{figure}

To better understand observability classes, consider the following scenario with a cup of coffee and some sugar on the table (Figure~\ref{fig:obs_attr}). 
Initially, the cup is far from the man, with no sugar. When he is not attentive, the robot adds some sugar to the cup and pushes it toward him. 
Once he is attentive again, he would notice that the cup is closer as its position is \textit{observable} to him.
But, without seeing the robot adding sugar, he cannot assess if there is sugar in the cup. Hence, we say the \textit{SugarInCup} attribute is \textit{inferrable}, 
and the human agent would know if they observed the robot act.

\section{Belief Updating}
An agent's belief can get updated in four ways as follows. 

\paragraph{When Acting}
If the agent executes an action, its belief is updated based on its effects. The state attributes appearing in the action's effect get updated with the new values, while the remaining attributes keep their same old values.

\paragraph{When Being an Observant} \textbf{(\textit{Action Observability.})}
An action executed by an agent is \textit{observed} by another agent \emph{co-present} throughout the action execution. Therefore, we state that if an agent observes an action getting executed, the agent will infer the \textit{inferrable} effects of the action.
Consequently, the agent immediately updates its belief state while each \textit{inferrable} attribute affected by the action receives a new value.  



However, when the human executes an action not observed by the robot, we keep it simple and consider that humans make only deterministic moves. The robot does not need to be co-present to get the effects of human actions. And, its belief state is always updated with both their \textit{inferrable} and \textit{observable} effects. More complex cases are part of our future work where the robot's beliefs can diverge, too. 

\paragraph{Via Situation Assessment}

The agent assesses the situation from its current location via spatial reasoning and its reference frame, i.e., the \textit{observability} model. Consequently, it updates its existing belief with the relevant ground truth. 

(Based on Definition~\ref{def:pssav}) We can always associate specific state attributes to \textit{places}. 
For example, suppose that being in the state $s_1$, the robot switches on the furnace placed in \textit{kitchen}, and this generates a new state $s_2$. Also, assume that
$f_{\textit{TurnOn}}
$ is $\observable$ (Definition~\ref{def:svof}).   
Then, the \text{place specific attribute function}, $f_{loc} : (...)$, w.r.t. the states
$s_1$ and 
$s_2$ can be expressed as, 
$f_{\textit{loc}} (f_{\textit{TurnOn}}^{s_1} )$ and $f_{\textit{loc}} (f_{\textit{TurnOn}}^{s_2} )$, respectively, and both map to $\textit{kitchen} \in Places$.

Consider the following scenario: The search progresses from $s_2$, along $s_1, s_2, ...$, such that the next action applicable in it is, the agent $(\varphi_h)$ \textit{moving} to the kitchen. This generates a new state $s_3$, and hence $f_{\textit{loc}} (f_{\textit{AgtAt}}(\varphi_h, s_3) )$ maps to $kitchen$, but so does  $f_{\textit{loc}} (f_{\textit{TurnOn}}^{s_3})$. 
In such cases, $\varphi_h$ assesses the \textit{status} of the furnace, i.e., the exact value of $f_{\textit{TurnOn}}^{s_3}$ in $s_3$, which is {\sc on}, and hence the human agent updates their belief.



%
%


Note that the robot is always aware of the ground reality hence, technically, the idea 
is effective only for the human agent. Here, the robot takes the human's perspective and performs spatial reasoning as per the human's frame of reference or their current location in the environment. 
The human's belief is updated w.r.t. what they can see as ground truth (mimicked by the robot).
This enables $\varphi_h$ to learn an \textit{observable} attribute's value achieved earlier by the robot.
%

The agent updates its belief with relevant ground truth learned via situation assessment, performed ``immediately'' (\textit{i.e.}, takes $0$ seconds) at each stage, before the execution of the next action. Technically, the robot performs situation assessment for the human and itself.

\paragraph{When Being Communicated}
The agent's belief is updated by learning the ground truth when another agent communicates an attribute-value pair. Communication occurs via a \textit{distinct set of actions}, modeled as a predefined communication protocol for a pair of sender-receiver agents. 



\section{Planning With Communication Actions}
For an agent to decide \textit{if}, \textit{when}, and \textit{what} to communicate to another agent, it first needs to establish a common protocol with that agent even before planning.  
In this work, we establish such communication protocols between each sender-receiver agents pair, where we use speech modality, and model explicit communication actions between two agents. 
However, we note that these actions are different from agents' (non-) primitive actions, and they are used only in special circumstances during planning and execution.

\subsection{Modeling Communication Actions} 
Next, we propose a generic communication action schema ($ca$) in this context. 
%
Suppose there exists a state-variable function,  $f_{svs}:(?g_1 (gr_1), ?g_2 (gr_2), ..., ?g_k (gr_k),\mathcal{S}) \rightarrow ?g_{k+1} (gr_{k+1})$, such that
for the current world state $s \in \mathcal{S}$, 
$f_{\textit{svs}}(g_{(1)l},g_{(2)m},...,g_{(k)n},s)$ maps to $q$.

The agent $\varphi_i$ can \textit{communicate} 
this attribute-value pair to $\varphi_j$ (\textit{via} the action $ca_{\varphi_i, \varphi_j}(f_{\textit{svs}}(...),q)$), if the following conditions (i.e., its preconditions) meet: 
(1) For $\varphi_i$, 
$f_{\textit{svs}}(g_{(1)l},g_{(2)m},...,g_{(k)n},B_{\varphi_i}^s) = q$, and (2) for $\varphi_j$,
$f_{\textit{svs}}(g_{(1)l},g_{(2)m},...,g_{(k)n},B_{\varphi_j}^s) = r$ s.t.
$r \neq q$.
Later, $\varphi_j$'s belief is updated, i.e., $f_{\textit{svs}}(g_{(1)l},g_{(2)m},...,g_{(k)n},B_{\varphi_j}^s) \leftarrow q$. 

%


\subsection{Planning with Reasoning on Human Mental State }
Now that we have defined and described all essential tools, we describe the new approach and explain how these tools are utilized by it. 
First, we note that we put an effort to make our main contributions to have a ``generic'' state-variable representation. Hence, we believe that they are not limited to only our intended architecture (HATP/EHDA) in this paper. We plan to substantiate this claim experimentally in future.

We only describe the \textit{main changes} made to enhance the underlying solver. 
Suppose an agent applies an action: First, the belief states of all agents co-present with this agent get updated with the action's inferrable effect. Later, the \textit{situation assessment} process is used for every other agent to assess the changes (captured via observable state attributes). As a result, their beliefs are updated by the observable effects if the required conditions are satisfied. We discussed these subroutines in detail in earlier sections.

\begin{figure*}[!ht]
    \centering
    \includegraphics[width=0.78\linewidth]{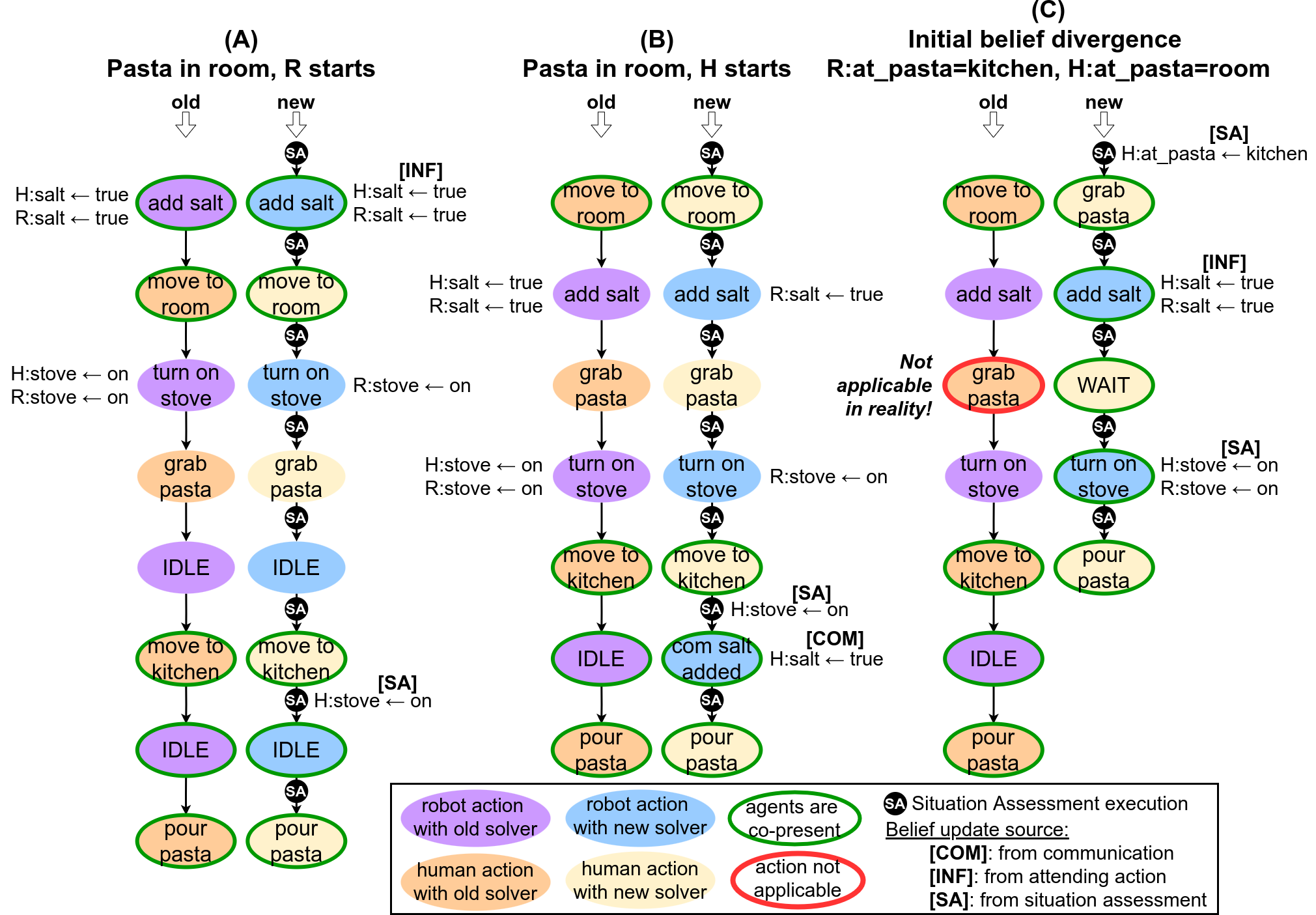}
    \caption{
    Shows the plans obtained in three scenarios: Each scenario presents two plans --- on the left, obtained by the old solver, and on the right, obtained via the proposed solver. The latter depicts more realistic and appropriate belief updates focusing on two attributes.
    Case~(A): the plans are the same, but the updates in the human belief are more realistic with our approach. Case~(B): the human has no certainty on {\em SaltInPot}, while ours decides to communicate to remove the ambiguity. And, Case~(C): the initial belief divergence induces an ``invalid'' plan, but our approach predicts that the human agent will \textit{assess the situation} and update their belief with all the facts without being communicated. (Other minor details are outlined in the figure.)
    }
    \label{fig:scenarios}
\end{figure*}

\subsubsection{Communication in Planning}
The following essential changes in the existing algorithm support efficient planning with communication.
(Recall the sender-receiver pair discussed earlier.) The approach identifies whether the receiver's belief divergence is \textit{relevant}, and thus if the receiver's belief needs to be corrected.
If so, all state attributes where the receiver agent's belief differs w.r.t. the ground truth are identified. Then, we decide which of these state attributes and their exact values must be transmitted by the sender (\textit{i.e.}, the robot), supporting the idea of communicating minimally and avoiding verbosity. Once we determine that, the receiver's (\textit{i.e.}, human's) belief gets updated accordingly, ensuring that the remaining false knowledge of the receiver's updated belief is \textit{not effective}, which also means that the belief divergence is not relevant anymore.



\begin{itemize}
    \item \textbf{Relevant Belief Divergence.}
    At a given stage, if the human agent, based on their own belief, can perform a set of actions that differs from 
    the one 
    the human could perform w.r.t. to the robot's belief (or the ground reality), then we say that such divergence in their belief is \textit{relevant}.
    A divergence is also declared as \textit{relevant} if an action has different effects w.r.t the other agent's belief.
    However, currently, our reasoning system is myopic: We do not evaluate in a principled way the impact of \textit{different} actions that humans could execute (based on their wrong belief), or of \textit{different} effects, with their overall positive and detrimental effects on achieving the joint task. At this stage, it simply ``decides'' to align the agent's beliefs using communication actions whenever the belief divergence is relevant.
    However, a smarter approach would probably be to analyze the history of plan trace(s) or agents' future actions to decide whether their beliefs should get aligned or not, or for that matter, how much to communicate, but it is currently out of the scope of this work. 
    
    \item \textbf{Communicate Only The Required Facts.}
    It ``decides'' the \textit{key} changes in the human's belief to be made such that the relevance of their updated belief (might still be diverging) is non-effective. 
    The subroutine that decides needed 
    communication to be made appears in the robot's executable policy; is described in the following steps:
    \begin{enumerate}
        \item 
        \textit{Store} each attribute and its value if the attribute's value differs in the human's belief from the robot's belief. 
    
        \item For each stored attribute-value pair, \textit{build} a communication action $ca_{\varphi_r, \varphi_h}(...)$ based on the schema described earlier. They are considered equally costly.  
        \item 
        At a given planning stage, \textit{e.g.}, the ground truth $s_i$, follow the Breadth-First Search ordering. 
        The \textit{source} is $B_{\varphi_h}^{s_i}$, and each $ca_{\varphi_r, \varphi_h}(...)$ changes and aligns \textit{exactly} one attribute while its updated value satisfies the ground truth. 
        Applying $ca_{\varphi_r, \varphi_h}(...)$ generates a new belief state following state transition rules, i.e., $B_{\varphi_h}^{s_i,1} = \gamma(B_{\varphi_h}^{s_i}, ca_{\varphi_r, \varphi_h}(...))$.
        It continues until the first (updated) belief is \textit{selected} to expand s.t. its remaining divergence is ineffective. The actions used from the root until the current belief state are \textit{retrieved}.
    \end{enumerate}
    Once the above subroutine finishes, the retrieved action set $\mathit{CA} = \{ca^{i}_{\varphi_r,\varphi_h}(...)\}$ is utilized for belief alignment. 
\end{itemize}


Agents may begin with different beliefs. Situation assessment (SA) can update human belief state, but humans cannot know inferable facts. So, we redefine Definition~\ref{def:hatp-joint-sol-plan} to make it sound w.r.t. our new planning approach. 
Assume at each step ($t=0,1,2,...$), humans perform SA, while the robot executes each communication action $ca \in \mathit{CA}$, such that the human's belief state \textit{updates immediately} (takes $0$ seconds). 
Later, following the updated beliefs, the human's or robot's regular primitive action is executed, and the effects only appear in the next step. Hence, the other two types of belief updates happen post the action execution. And, continues.

\section{Empirical Evaluation}

Before we describe the two domains used for the experiments and the results to compare the old and new approaches, we first make a \textit{high-level} distinction between the existing and new approaches. 
In principle, the existing solver can handle agents' individual belief divergences, but ``not during planning'' unlike our approach. To achieve that, the existing solver can use a cumbersome technique that intervenes by updating the (collaborative) task networks of the joint task model/specifications while using triggers.


\paragraph{Cooking Pasta Domain}
Suppose a stove and salt are available in \textit{Kitchen} ($\textit{Places}$), and the pasta is either in \textit{Kitchen} or \textit{Room} -- they are adjacent. The agents have different roles and can only operate in the two places. Robot \textit{adds} salt to the pot and \textit{turns-on} the stove. Human \textit{grabs} the pasta and \textit{pours} it into the pot. 
The human can add pasta, but only after salt gets added to the pot and the stove is {\sc on}.

Focus on the following two attributes: For a given state, $s_i \in \mathcal{S}$, $f_{\textit{SaltInPot}}^{s_i} \in \{\textit{true, false}\}$ and $f_{\textit{stove}}^{s_i} \in \{\textit{on, off}\}$ such that only 
$f_{\textit{SaltInPot}}^{s_i}$
is \textit{inferrable}, all others belong to $\observable$.

\paragraph{Preparing Box Domain}
A box filled with a fixed number of balls and with a sticker pasted on is considered prepared and needs to be sent. Both the agents can \textit{fill} the box with balls from a bucket, while only the robot can \textit{paste} a sticker, and only the human can \textit{send} the box. The bucket can run out of balls, so when there is only one ball left, the human \textit{moves} to another room to \textit{grab} more balls and \textit{refill} it. A sticker pasted on the box is \textit{observable}, while the number of balls in the box is \textit{inferrable}. 
Other attributes belong to $\observable$.   

\subsection{Experiments}
Qualitative analysis in the first domain mentions \textit{subtleties} the old solver overlooks and how ours is aware of them, updates belief, and effectively manages divergences. 
Then, quantitative studies compare the solvers in both domains.

\subsubsection{Qualitative Analysis}
Consider the first domain. We discuss the plans obtained in three different scenarios as shown in Figure~\ref{fig:scenarios}. Assume that both the agents are in \textit{Kitchen} and the pasta is in \textit{Room}. Scenario~(A) captures the agents' plans when the robot starts while Scenario~(B) shows when the human starts. Let us change the scenario: Suppose, in hindsight, the pasta is moved to \textit{kitchen} such that the robot knows it, while the human still has the old belief. For this, Scenario~(C) captures the plans obtained when the human starts. 
\begin{itemize}
    \item \textbf{Scenario~(A):} 
    Although the solvers generate similar plans, they update the human's belief differently if and when the human and robot are co-present. E.g., turning on the stove ideally (realistically) does not affect the human's mental state, which is not the case for the old solver. It considers agents omniscient, so the human knows everything immediately once achieved. 
    Our solver predicts it when the human returns to the kitchen and assesses that the stove is {\sc on}, and their belief gets updated.
    \item \textbf{Scenario~(B):} Human leaves the kitchen. Practically, they are unaware of the changes achieved in the environment by the robot's actions: {\em turn-on} \& {\em add-salt}. The old solver generated a similar plan as in {\em Sce.~(A)} where the human already knew that the stove was {\sc on}, and salt was added.
    However, an appropriate situation assessment and inference based on action observability together guarantee that, when returning to the kitchen, the human assesses that the stove is {\sc ON}. Moreover, the robot predicts that the human cannot know the \textit{SaltInPot} fact and that it is a relevant divergence that needs to be handled via communication. 
    \item
    \textbf{Scenario~(C):}
    With the old solver, the human moves to \textit{Room} to \textit{grab} the pasta, but in reality, being an illegal action w.r.t. the robot's belief or the ground truth, it \textit{fails}. 
    In our case, the human agent assesses the environment to update their belief state, knowing the pasta is in the kitchen. 
    Hence, no communication is required.
\end{itemize}

\begin{table}
    \begin{adjustbox}{width=0.83\columnwidth,center}
    \begin{tabular}{@{}c|r r r|r r@{}}
        \multirow{2}{*}{
        \textbf{Domain}} & \multicolumn{3}{c|}{\textbf{\textit{Old Solver}}} & \multicolumn{2}{c}{\textbf{\textit{Our Solver}}}
        \\
        & \multicolumn{1}{c}{\textit{S}} & \multicolumn{1}{c}{\textit{NA}} & \multicolumn{1}{c|}{\textit{IDL}} & \multicolumn{1}{c}{\textit{S}} & \multicolumn{1}{c}{\textit{Com}} 
        \\ \cline{1-6}
        \textit{Cooking} & 18.6\% & 77.0\% & 23.0\% & 100\% & 54.9\%\\
        \textit{Box} & 25.0\% & 83.3\% & 16.7\% & 100\% & 68.8\%\\
        \hline
        \textbf{Average} & 21.8\% & 80.2\% & 19.9\% & 100\% & 61.9\%\\
    \end{tabular}
    \end{adjustbox}
    \caption
    {
    \label{tab:q_results}
    In each domain: for the \textit{old solver}, the success rate (\textit{S}), the ratio of failed plans due to a non-applicable action (\textit{NA}), and the ratio of failed plans due to an inactivity deadlock case (\textit{IDL}), while for \textit{our solver}, the success rate (\textit{S}) and the ratio of plans including a communication action (\textit{Com}).
    }
\end{table}

\subsubsection{Quantitative Studies}
The HATP/EHDA framework's current solver and ours are tested and compared on two domains in Table~\ref{tab:q_results}. We consider (in box domain) \textit{three} boxes to be prepared and sent. While in both, we generated 512 different initial states, including 448 (87.5\%) with divergent initial agents' beliefs and 64 states where both agent beliefs are fully aligned initially. Overall, 2048 plans were generated.

With the old approach, a planning failure occurs due to: 
(a) an action of a plan not applicable in another agent's belief state, including the ground truth; 
(b) if an inactivity deadlock occurs, which is assumed to be the case after a succession of at least four \textit{WAIT} and (or) \textit{IDLE} actions. 
A deadlock occurs when the human has a belief divergence and waits for a never-happening robot's action, e.g., waiting for \textit{add salt}, but \textit{SaltInPot} is already achieved.
For the old solver, the success rate (\textit{S}), the \textit{ratios} of the number of failed plans due to a inapplicable action (\textit{NA}) and an inactivity deadlock (\textit{IDL}) appear in the table.
For ours, the success rate (\textit{S}) is shown and the \textit{ratio} of successful plans including a communication action is presented under \textit{Com}.

As the existing solver does not handle belief divergence in planning, the applicability of actions was never an issue w.r.t. another agent's belief. Therefore, if the \textit{IDL} case occurs, it is understood that the task specification is erroneous w.r.t. the old problem specification. 

Our solver always finds legal plans, and on average, \textit{approx.} 62\% of them use \textit{communication}.
Thus, the robot doesn't need to communicate systematically as assessing situations handles a major part of the divergences (87.5\% of the scenarios have divergent beliefs initially).
For the old solver, if no initial belief divergence exists, it always finds a legal plan, considering the agents omniscient. 
E.g., Fig.~\ref{fig:scenarios}(A). However, sometimes, this causes problems in practice.
Scenarios beginning with distinct beliefs induce actions often not applicable in another agent's belief state (or the ground truth), evident by the (average) 21.8\% success rate.

\section{Discussion}
Formalizing run-time observability conventions is crucial for planning as ignoring them may lead to problems in practice. 
E.g., Fig.~\ref{fig:scenarios}(B) showing human knowing \textit{SaltInPot}, is ambiguous in reality. We describe a far more realistic way to estimate the evolution of human belief during planning, using the situation assessment discussed in \textit{Qualitative Analysis}. Explicit reasoning on the human mental state detects and prevents ambiguous situations with communication while also prohibiting the robot from being
too verbose,
as shown in \textit{Quantitative Studies}. 
Compared to the old process, this produces \textit{consistent} and more \textit{robust} plans overall.

However, our approach does not \textit{refute} something believed by an agent through situation assessment without assessing its exact true value. 
E.g., for some $s_i \in \mathcal{S}$, if the human 
wrongly believes that the pasta is in \textit{Kitchen}. The situation assessment does not help refute this, while the human is in \textit{Kitchen}. 
The reason is that $f_{\textit{PastaNotInKitchen}}^{s_i}(...)$ is not modeled explicitly as an attribute. 
Such issues do not affect the solver's \textit{completeness} as far as the situation assessment is concerned. Moreover, our solver handles them as a relevant divergence to be aligned. 
Thus, the human is just communicated with correct updates.

\section{Summary} 
Building on earlier work, we presented a new method to model execution-time observability conventions appropriate for HRI and to use them to estimate the evolution of the human mental state. It is based abstractly on situation assessment and action observability criteria. 
A new planning approach is described, utilizing this better estimation of human mental state to plan for more robust and consistent human-robot joint activities such that a relevant belief divergence is tackled by explicitly modeled communication actions.

\bibliography{bib.bib}


\end{document}